\pdfoutput=1

\documentclass[11pt]{article}

\usepackage{emnlp2021}

\usepackage{times}
\usepackage{latexsym}

\usepackage[T1]{fontenc}

\usepackage[utf8]{inputenc}

\usepackage{microtype}
\usepackage{graphicx}
\usepackage{amsmath}
\usepackage{amssymb}
\usepackage{amsthm}
\usepackage{booktabs}
\usepackage{algorithm}
\usepackage{algorithmic}
\usepackage{multirow} 
\usepackage{color} 
\usepackage{subfigure} 
\title{Adversarial Mixing Policy for Relaxing Locally Linear Constraints in \textit{Mixup}}

 \author{Guang Liu,  Yuzhao Mao,  \\  \bf{Hailong Huang,  Weiguo Gao, \and Xuan Li} \\
         	PingAn Life Insurance of China\\
		\url{https://github.com/PAI-SmallIsAllYourNeed/Mixup-AMP}
          }

\begin{document}
\maketitle
\begin{abstract}

Mixup is a recent regularizer for current deep classification networks. Through training a neural network on convex combinations of pairs of examples and their labels, it imposes locally linear constraints on the model’s input space. However, such strict linear constraints often lead to under-fitting which degrades the effects of regularization. Noticeably, this issue is getting more serious when the resource is extremely limited. To address these issues, we propose the Adversarial Mixing Policy (AMP), organized in a ``min-max-rand'' formulation, to relax the Locally Linear Constraints in Mixup. Specifically, AMP adds a small adversarial perturbation to the mixing coefficients rather than the examples. Thus, slight non-linearity is injected in-between the synthetic examples and synthetic labels. By training on these data, the deep networks are further regularized, and thus achieve a lower predictive error rate. Experiments on five text classification benchmarks and five backbone models have empirically shown that our methods reduce the error rate over Mixup variants in a significant margin (up to 31.3\%), especially in low-resource conditions (up to 17.5\%). 
\end{abstract}

\section{Introduction}

Deep classification models have achieved impressive results in both images~\cite{he2016deep,dosovitskiy2020image} and language processing~\cite{devlin2018bert,kim2014convolutional,wang2016attention}. One of the most significant challenges to train a deep model is the great efforts and costs to collect large-scale labels. Without sufficient labels, the deep networks tend to generalize poorly, leading to unsatisfactory performance. Thus, the regularization techniques under augmentation schema, which generate labeled data to regularize models~\cite{hernandez2018data}, are widely explored~\cite{wei2019eda,DBLP:conf/dasfaa/LiuHMGLS21}. 




%


 \textit{Mixup}~\cite{zhang2017mixup} is an effective regularizer under the augmentation schema. In recent years, topics related to \textit{Mixup} have warranted serious attention~\cite{lee2020adversarial,xu2020adversarial,verma2019manifold,archambault2019mixup,berthelot2019mixmatch,berthelot2019remixmatch,beckham2019adversarial,mao2019virtual,zhu2020automix}. The core idea of \textit{Mixup} is to generate synthetic training data via a mixing policy, which convex combines a pair of examples and its labels. Through training on these data, the classification networks will be regularized to reach higher performance. Unlike conventional regularizers~\cite{srivastava2014dropout,hanson1988comparing,ioffe2015batch}, \textit{Mixup} imposes a kind of locally linear constraint~\cite{zhang2017mixup,guo2019mixup} on the model’s input space. 
 
 However, vanilla \textit{Mixup} often suffers from under-fitting due to the ambiguous data~\cite{guo2019mixup,guo2020nonlinear,mai2021metamixup} generated under the strict locally linear constraints. To alleviate the under-fitting, ~\cite{guo2020nonlinear} uses extra parameters to project the inputs and labels into a high dimensional space to properly separate the data.~\cite{guo2019mixup,mai2021metamixup} use auxiliary networks to learn the mixing policy in a data-driven way to avoid the generation of ambiguous data. Although existing works effectively reduce the under-fitting, they have limitations to properly regularization networks. Current networks are prone to be over-fitting when adding the extra parameters. Eventually, these methods degrade the effects of regularization. The conflicts between over-fitting and under-fitting get more serious when the labeled resources are rare or hard to obtain. Besides, the methods with auxiliary networks usually have difficulties in integrating with other \textit{Mixup} variants. More importantly, \textit{Mixup} works well in most cases~\cite{guo2019mixup}. Adding too much non-linearity into \textit{Mixup} will sacrifice the majority of synthetic data that can regularize the networks under locally linear constraints. So, the locally linear constraints in \textit{Mixup} only need to be slightly relaxed. 

In this paper, we propose the Adversarial Mixing Policy (\textit{\textbf{AMP}}) to overcome these limitations. 
We modify the adversarial training~\cite{goodfellow2014explaining}, which relaxes the linear nature of the network without any extra parameters or auxiliary networks, to relax the Locally Linear Constraints in \textit{Mixup}. Inspired by the ``min-max'' formulation of adversarial training, we formulate our method as a form of ``min-max-rand'' regularization. Specifically, the ``rand'' operation randomly samples a mixing coefficient as in vanilla \textit{Mixup} to generate synthetic example and label. Then, the ``max'' operation calculates the perturbation of the mixing coefficient and applies it. 
Note that the updated mixing coefficient is only used to re-synthetic example, keeping the synthetic label unchanged. Thus, slight non-linearity is injected in-between the synthetic example and label. Finally, the ``min'' operation minimizes the training loss over the non-linearly generated example-label pairs. In summary, we highlight the following contributions: 
\begin{itemize}
    \item We propose an Adversarial Mixing Policy (AMP) to relax the Locally Linear Constraints (LLC) in \textit{Mixup} without any auxiliary networks. It can be seamlessly integrated into other \textit{Mixup} variants for its simplicity.
    \item To the best of our knowledge, this is the first exploration of the application of adversarial perturbation to the mixing coefficient in \textit{Mixup}. 
    
    \item We analyze our proposed method with extensive experiments and show that our AMP improves the performance of two \textit{Mixup} variants on various settings and outperforms the non-linear \textit{Mixup} in terms of error rate.
\end{itemize}

\section{Background}
\subsection{Linear nature of the networks}
Let $(x;y)$ be a sample in the training data, where $x$ denotes the input and $y$ the corresponding label. Deep networks learns a mapping function from $x$ to $y$, which is:
\begin{equation}\label{eq21}
f(x) = y' \rightarrow y\,.
\end{equation}
Here, $y'$ is the output of the networks, $\rightarrow$ represents the learning process. The linear nature of networks can be interpreted as that a small change in the input will lead to a change of model output:
\begin{equation}\label{eq223}
f(x+\nabla x) = y'+ \nabla y\,.
\end{equation}
Here, $\nabla x$ is a small perturbation of $x$, and $\nabla y$ is the changing of output caused by the injection of $\nabla x$. This linearity causes the networks vulnerable to adversarial attacks~\cite{goodfellow2014explaining}.

\subsection{Relax the linear nature}
To relax the linear nature of the networks, adversarial training~\cite{goodfellow2014explaining} forces the networks to learn the following mapping function,
\begin{equation}\label{eq22}
    f(x+ \nabla x) = y' \rightarrow y\,,
\end{equation}
where $\nabla x$ is an small adversarial perturbation. Such kind of training can effectively relax the linearity of networks and improve the robustness of deep networks. However, there exists a trade-off between model robustness(Equation.~\ref{eq22}) and  generalization(Equation.~\ref{eq21})\cite{tsipras2018robustness}.


\subsection{Locally linear constraints in \textit{Mixup}}
Mixup can be formulated as follows,
\begin{align}
& f(m_x(\lambda))=y' \rightarrow m_y(\lambda) \,,\\
& m_{x}(\lambda) = x_1 \cdot \lambda + x_2 \cdot (1-\lambda)\,,\\
& m_{y}(\lambda) = y_1 \cdot \lambda + y_2 \cdot (1-\lambda)\,,
\end{align}
where $\lambda\in[0,1]$ is the mixing coefficient. $m$ is the mixing policy. $(x_1;y_1)$ and $(x_2;y_2)$ are a pair of examples from the original training data. 
By training on synthetic data, $m_{x}(\lambda)$ and $m_{y}(\lambda)$, \textit{Mixup}~\cite{zhang2017mixup,verma2019manifold} imposes the Locally Linear Constraints on the input space of networks. Different from Eq.~\ref{eq223}, this linearity can be formulated as follow,
\begin{equation}\label{eq23}
    f(m_x(\lambda +\nabla\lambda))= y'+\nabla y \rightarrow m_y(\lambda+\nabla\lambda)\,.
\end{equation}
Here, the $\nabla \lambda$ is a small change in $\lambda$. We can observe that the output of the networks is changed accordingly. That is similar to the form of the linear nature of networks. Under these settings, the small change in $\lambda$ often leads to an undesirable change of output. Eventually, these strict linear constraints lead to under-fitting that degrades the regularization effects~\cite{guo2019mixup,guo2020nonlinear}.  
\subsection{Why relaxing locally linear constraints}
Relaxing the strict linear constraints in \textit{Mixup} can alleviate the under-fitting and therefore improve the regularization effects~\cite{guo2020nonlinear}. The under-fitting happens when the synthetic data is corrupted or ambiguous for the network. So, if we can make the networks compatible with such data, like the soft margin~\cite{suykens1999least}, the under-fitting will be eased. Furthermore, such a technique is best realized the relaxing without extra parameters. Inspired by the adversarial training (Eq.~\ref{eq22}), we hypothesize that injecting slight non-linearity into \textit{Mixup} can relax its constraints without extra parameters 
as follow,
\begin{equation}\label{eq24}
f(m_x(\lambda+\nabla \lambda))=y' \rightarrow m_y(\lambda)\,,
\end{equation}
where $\nabla \lambda$ is an adversarial perturbation injected to the original mixing coefficient $\lambda$. 

\section{Methodology}
\begin{figure*}[ht]
	\centering
	\includegraphics[scale=1]{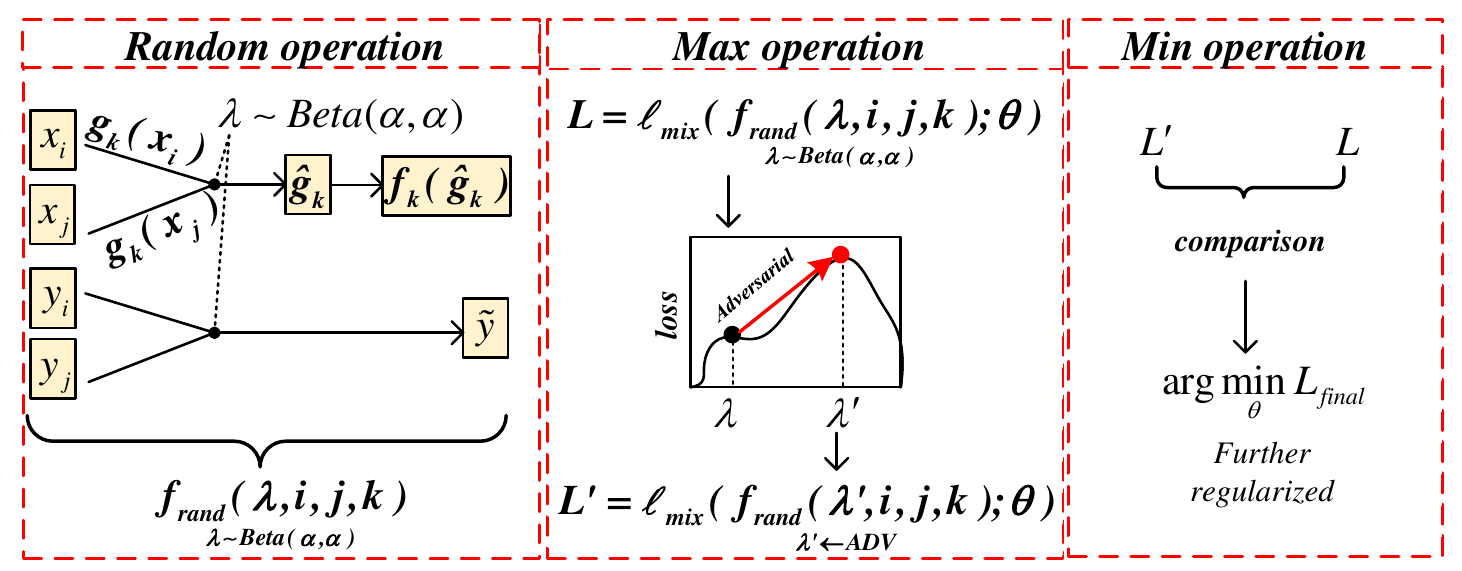}
	\caption{The major operations of Adversarial Mixing Policy (\textit{AMP}). \label{fig:structure}}
\end{figure*}

As shown in Figure~\ref{fig:structure}, Adversarial Mixing Policy (\textbf{AMP}) consists of three operations: \textbf{Rand}, \textbf{Max} and \textbf{Min}. Rand Operation (RandOp) generates the synthetic data by interpolating pairs of training examples and their labels with a random mixing coefficient $\lambda$. Max Operation (MaxOp) injects a small adversarial perturbation into the $\lambda$ to re-synthesize the example and keeps the synthetic label unchanged. This operation injects slight non-linearity into the synthetic data. Min Operation (MinOp) minimizes the losses of these data. Additionally, we use a simple comparison to eliminate the influence caused by the scaling of gradients.

\subsection{Method formulation}
Given a training set $D=\{x_i,y_i\}$ of texts, in which each sample includes a sequence of words $x_i$ and a label $y_i$. A classification model encodes the text into a hidden state and predicts the category of text. \textit{Mixup}'s objective is to generate interpolated sample $\hat{g_k}$ and label $\hat{y}$ by randomly linear interpolation with ratio $\lambda$ applied on a data pair$(x_i;y_i)$ and $(x_j;y_j)$. Our method aims to project a perturbation $\nabla\lambda$ into $\lambda$ to maximize the loss on interpolated data. Then, it minimizes the maximized loss. Inspired by adversarial training, we formulate this problem as a \textit{\textbf{min-max-rand}} optimization problem,
\begin{equation}
\min_\theta\mathbb{E}_{\hat{D}}\max_{|\nabla \lambda|\leq\varepsilon}\ell_{mix}(\mathop{f_{rand}(\lambda+\nabla \lambda,i,j,k)}_{\lambda\sim Beta(\alpha,\alpha)};\theta)\,.
\end{equation}
Here, $\hat{D}=\{\hat{g_k}_i,\hat{y}_i\}$ is the synthetic data set generated by $f_{rand}(\lambda,i,j)$, $\nabla \lambda$ is the adversarial perturbation of $\lambda$, $\varepsilon$ is the maximum step size, $\ell_{mix}(\ast)$ is the \textit{Mixup} loss function, $f_{rand}(\ast)$ represent the random interpolation of data and labels, $\lambda$ is the random mixing coefficient sampled from a $Beta$ distribution with $\alpha$ parameters, $i$ and $j$ are the randomly sampled data indexes in $D$, $k$ is the mixed layer.
\subsection{Rand operation}
Rand Operation (RandOp) is identical to \textit{Mixup}~\cite{zhang2017mixup}. It aims to generate random interpolated data between two categories. Specifically, it generates synthetic labeled data by linearly interpolating pairs of training examples as well as their corresponding labels. For a data pair $(x_i; y_i)$ and $(x_j; y_j)$, $x$ denotes the examples and $y$ the one-hot encoding of the corresponding labels. Consider a model $f(x)=f_k(g_k(x))$, $g_k$ denotes the part of the model mapping the input data to the hidden state at layer $k$, and $f_k$ denotes the part mapping such hidden state to the output of $f(x)$. The synthetic data is generated as follows,
\begin{align}
\lambda \sim  &Beta(\alpha,\alpha)\,,\label{eq1}\\
\hat{g_k}= &g_{k}(x_i) \cdot\lambda +  g_{k}(x_j)\cdot(1-\lambda)\,,\label{eq2}\\
\hat{y}= &y_i\cdot \lambda  +  y_j\cdot (1-\lambda)\label{eq3}\,,
\end{align}
where $\lambda$ is the mixing coefficient for the data pair, $\alpha$ indicates the hyper-parameter of $Beta$ distribution, $\hat{g_k}$ is the synthetic hidden state. For efficient computation, the mixing happens by randomly picking one sample and then pairs it up with another sample drawn from the same mini-batch~\cite{zhang2017mixup}. Here, the sample is obtained randomly. To simplify, we reformulate the random interpolation $f_{rand}(\ast)$ as follow, 
\begin{equation}
(f_k(\hat{g_k}),\hat{y}):=\mathop{f_{rand}( \lambda,i,j,k)}_{\lambda\sim Beta(\alpha,\alpha)}\,.
\end{equation}
Here, $f_{rand}(\ast)$ takes the results of Equation~\ref{eq1}-~\ref{eq3} as input, outputs the model predictions $f_{k}(\hat{g_k})$ and the label $\hat{y}$. The model trained on the generated data tends to reduce the volatility of prediction on these data. Then, the model will generalize better on unseen data.

\subsection{Max operation}
Max operation (MaxOp) injects a small adversarial perturbation to inject slight non-linearity between the synthetic example and synthetic label. It means that the generated synthetic data will not strictly follow the Locally Linear Constraints in \textit{Mixup}. To achieve this, we propose an algorithm, which is similar to the Fast Gradient Sign Method (FGSM)~\cite{goodfellow2014explaining}, to inject an adversarial perturbation to the $\lambda$. It calculates the gradient of $\lambda$ in the gradient ascend direction,
\begin{equation}
\max_{|\nabla \lambda|\leq\varepsilon}\ell_{mix}(\mathop{f_{rand}(\lambda+\nabla\lambda,i,j,k)}_{\lambda\sim Beta(\alpha,\alpha)};\theta)\,,
\end{equation}
where the $\nabla \lambda$ is the gradients of $\lambda$ on gradient ascent direction, $\varepsilon$ is the step size. Different from the FGSM~\cite{goodfellow2014explaining}, we add a small perturbation on $\lambda$ instead of the input. Besides, the $\lambda$ is a scalar, we can get the adversarial direction and strength directly. So, there is no need to perform the normalization on $\nabla \lambda$.
\begin{equation}
\lambda' = \lambda+\varepsilon\cdot \nabla\lambda\,,
\end{equation}
where $\lambda'$ is the slight hardness version of mix coefficient, $\varepsilon$ is the step size, $\nabla \lambda$ is the clipped ($\leq1$) gradient of $\lambda$. The perturbation is the gradient in the adversarial direction. We calculate the gradient of $\lambda$ as follow,
\begin{equation}
\nabla\lambda =  \frac{\partial \mathcal{L}}{\partial \lambda}\,.
\end{equation}
Here, the \textit{Mixup} loss $\mathcal{L}$ is calculated by interpolation of losses on pair of labels~\cite{zhang2017mixup,verma2019manifold} as follow,
\begin{equation}
\begin{split}
\mathcal{L}=&\ell_{mix} (\mathop{f_{rand}(\lambda,i,j,k)}_{\lambda\sim Beta(\alpha,\alpha)};\theta)\\
 =&  \ell_{ce}(f_k(\hat{g_k}), y_i;\theta)\cdot \lambda  +\\
 & \ell_{ce}(f_k(\hat{g_k}), y_j;\theta)\cdot(1-\lambda)  \label{eq5}\,. 
\end{split}
\end{equation}
Here, $\mathcal{L}$ represents the loss of synthetic data generated under mixing coefficient $\lambda$, $\theta$ is the parameters of the model, $\ell_{mix}(\ast)$ is the \textit{Mixup} loss, $\ell_{ce}(\ast)$ represents the cross-entropy function. Notable that the step size of gradient $\varepsilon$ may lead to undesirable results that minimize the losses. So, we need to eliminate the influence caused by $\varepsilon$. 
\subsection{Min operation}
Min operation (MinOp) minimizes loss of constraints relaxed synthetic data as follow, 
\begin{equation}
\mathop{\arg\min}_{\theta} \mathcal{L}_{final} \,,
\end{equation}
where $ \mathcal{L}_{final}$ is the final loss. In addition, MinOp leans to minimize the larger loss in the previous two steps to eliminate the influence of the step size $\varepsilon$. Besides, this preference will help model learning from the one with larger loss to reduce the risk of under-fitting. We use a mask-based mechanism to realize the operation as follow,
\begin{equation}
    \mathcal{L}_{final}=\mathcal{L} \cdot (1-mask)+ \mathcal{L}' \cdot mask\,.
\end{equation}
Here, the $mask$ is used as a selector of losses. The comparison is carried out on losses before and after updated $\lambda$ in the synthetic example. The latter one $\mathcal{L}'$ is calculated as follow,
\begin{equation}
    \mathcal{L}'=\ell_{mix}(\mathop{f_{rand}( \lambda',i,j,k)}_{\lambda'\leftarrow ADV};\theta)\,.
\end{equation}
Here, $\lambda'$ is the mixing coefficient after injecting perturbation (we only inject the perturbation into mixing coefficient of input, as Eq.~\ref{eq24}),  $\mathcal{L}'$ is the \textit{Mixup} loss on synthetic example generated under $\lambda'$. Note that the $\lambda$ for the synthetic label is unchanged. $mask$ is calculated as follow,
\begin{equation}
mask = \left\{\begin{matrix}
 1& \delta_\mathcal{L} > 0\\ 
 0& \delta_\mathcal{L} \leq 0\,.
\end{matrix}\right.
\end{equation}
Here, the $mask$ is batch size vector, $\delta_\mathcal{L}$ is the direct comparison $\mathcal{L}'-\mathcal{L}$. By doing this, the proposed method achieves steady improvement under different settings of step size.   
\section{Experiments}
\subsection{Data}
We evaluate the proposed \textit{AMP} on five sentence classification benchmark datasets as used in~\cite{guo2019augmenting}. TREC is a question dataset which aims to categorize a question into six types~\cite{li2002learning}. MR is a movie review dataset aiming at classifying positive/negative reviews~\cite{pang2005seeing}. SST-1 is the Stanford Sentiment Treebank dataset with five sentiment categories: very positive, positive, neutral, negative, and very negative~\cite{socher2013recursive}. SST-2 is a binary label version of SST-1. SUBJ is a dataset aiming to judge a sentence to be subjective or objective~\cite{pang2004sentimental}. Table~\ref{tab:1} summarizes the statistical characteristics of the five datasets after prepossessing.
\begin{table}[ht]
\centering
\caption{The statistics of datasets. $c$ is the category number. $l$ is the average length. $V$ is the vocabulary size. $N$ is the size of the training set. $T$ is the size of the testing set. $CV$ denotes the 10-fold cross-validation.\label{tab:1}}
\setlength{\tabcolsep}{2.8mm}
\begin{tabular}{cccccc}
\hline
Data  & $c$ & $l$  &  $V$  & $N$     & $T$ \\ \hline
TREC  & 6 & 10 &9592 & 5952  & 500  \\
SST-1 & 5 & 18 &17836& 11855 & 2210 \\
SST-2 & 2 & 19 &16185& 9613  & 1821 \\
SUBJ  & 2 & 23 &21323& 10000 & CV   \\
MR    & 2 & 20 &18765& 10662 & CV   \\ \hline
\end{tabular}
\end{table}
\begin{table*}[t]
\centering
\caption{The results of our \textit{AMP} method compared with two recent \textit{Mixup} methods on five different datasets under five different classification models. For a fair comparison, we re-implement the \textit{Mixup} baselines based on backbone models. The results may not the same as the results in~\cite{guo2019augmenting,sun2020mixup}. $RP$ indicates the relative improvement.$^\dag$ indicates the results are cited from ~\cite{guo2020nonlinear}.\label{tab:2}}
\setlength{\tabcolsep}{4mm}
\small 
\begin{tabular}{ccccccc}
\toprule
Model                                    & \textit{Mixup}   & TREC(\%)       & SST-1(\%)      & SST-2(\%)      & SUBJ(\%)       & MR(\%)         \\ \midrule
\multicolumn{1}{c}{\multirow{7}{*}{\small{$RNN_{rand}$}}} & \textit{w/o}   & 11.3±1.48                                & 63.7±3.00                              & 18.0±0.85                               & 10.7±0.57                               & 24.9±1.11 \\
\multicolumn{1}{c}{}                     & \textit{Sent}         & 10.5±1.16                                & 55.8±0.75                              & 16.6±0.38                               & 10.3±0.55                               & 24.2±0.72 \\
\multicolumn{1}{c}{}                     & \textit{Sent(\textbf{our})}& 9.8±0.73  & \textbf{55.0±0.37}             & 15.9±0.43 &  10.0±0.78 &  23.6±0.65 \\
\multicolumn{1}{c}{}                     & \textit{RP}(\%)             & $6.7_{\uparrow}$  & $1.4_{\uparrow}$            & $4.2_{\uparrow}$  & $2.9_{\uparrow}$  & $2.5_{\uparrow}$  \\
\multicolumn{1}{c}{}                     & \textit{Word}            & 9.8±0.86                                & 55.9±0.62                              & 16.1±0.62                               & 9.4±0.77                               & 23.6±0.75 \\
\multicolumn{1}{c}{}                     & \textit{Word(\textbf{our})}   &  \textbf{9.5±0.84}  & 55.6±0.67&\textbf{15.3±0.43} & \textbf{8.8±0.48} &  \textbf{22.7±0.96} \\
\multicolumn{1}{c}{}                     & \textit{RP}(\%)   & $3.1_{\uparrow}$  & $0.5_{\uparrow}$ & $5.0_{\uparrow}$  & $6.4_{\uparrow}$  & $3.8_{\uparrow}$ \\ \midrule
\multicolumn{1}{c}{\multirow{7}{*}{\small{$RNN_{glove}$}}}   & \textit{w/o}& 8.3±0.47                                & 56.6±0.30                              & 13.0±0.51                               & 6.1±0.76                               & 18.5±0.97 \\
                                         & \textit{Sent}         & 6.9±0.55                                & 48.1±0.37                              & 12.1±0.61                               & 6.0±0.69                               & 18.1±0.95 \\
                                         & \textit{Sent(\textbf{our})}&  6.7±0.27 & \textbf{48.0±0.45} & 11.5±0.31 & 5.8±0.79 &   17.8±0.98 \\
                                         & \textit{RP}(\%)             & $2.9_{\uparrow}$  & $0.2_{\uparrow}$  & $5.0_{\uparrow}$& $3.3_{\uparrow}$  &  $1.7_{\uparrow}$  \\
                                         & \textit{Word}            & \textbf{6.5±0.45}                                & 48.6±0.33                              & 11.8±0.34                               & 5.5±0.73                               & 17.8±0.87 \\
                                         & \textit{Word(\textbf{our})}   &  6.6±0.52  & \textbf{48.0±0.66} &  \textbf{11.1±0.42} &  \textbf{5.2±0.72} &   \textbf{17.5±0.91} \\ 
                                         & \textit{RP}(\%)           & $1.5_{\downarrow}$   & $1.2_{\uparrow}$  & $5.9_{\uparrow}$  &  $5.5_{\uparrow}$  &  $1.7_{\uparrow}$  \\ \midrule
\multirow{7}{*}{\small{$CNN_{rand}$}}                & \textit{w/o}        & 8.8±0.86                                & 63.2±0.54                              & \textbf{17.6±0.52}                               & 9.5±0.64                               & 24.2±1.39 \\

                                         & \textit{Sent}        & 8.3±0.63                                & 58.1±0.48                              & 19.9±0.32                               & 9.5±0.52                               & 25.1±0.91 \\
                                         & \textit{Sent(\textbf{our})}& \textbf{8.1±0.71}  &  57.9±0.51 & 19.9±0.51 &   9.4±0.45               & 25.1±0.93 \\
                                         & \textit{RP}(\%)         & $2.4_{\uparrow}$   & $0.5_{\uparrow}$  &${_\rightarrow}$  &  $1.1_{\uparrow}$ &$_\rightarrow$  \\
                                         & \textit{Word}            & 8.3±0.71                                & 58.0±0.55                              & 19.4±0.22                               & 9.7±0.57                               & 24.6±0.78 \\
                                         & \textit{Word(\textbf{our})}   &  8.4±0.92  &  \textbf{57.5±0.50} &  19.2±0.53 &  \textbf{9.2±0.68} &  \textbf{24.1±0.98} \\
                                         & \textit{RP}(\%)           & $1.2_{\downarrow} $   & $1.0_{\uparrow}$  & $1.0_{\uparrow}$  & $5.2_{\uparrow}$  &  $2.0_{\uparrow}$  \\
                                          \midrule
\multirow{7}{*}{\small{$CNN_{glove}$}}               & \textit{w/o}       & 7.9±0.12                                & 57.5±0.50                              & 13.1±0.49                               & 5.6±0.36                               & 20.2±0.60 \\
& \textit{\textit{Non-linear}}        & 5.3±0.29$^\dag$                                & 50.7±0.42$^\dag$                               & \textbf{11.4±0.29}$^\dag$                                & 6.1±0.19$^\dag$                                & \textbf{16.6±0.36}$^\dag$  \\
                                         & Sent         & 6.7±0.23                                & 51.4±0.23                              & 12.8±0.35                               & 5.1±0.34                              & 19.4±0.56 \\
                                         & \textit{Sent(\textbf{our})}&  \textbf{4.6±0.33} &  50.6±0.40&  11.7±0.25 & \textbf{5.1±0.62}&  17.4±0.69 \\
                                         & \textit{RP}(\%)           & $31.3_{\uparrow} $ & $1.6_{\uparrow}$& $8.6_{\uparrow}$  & $_\rightarrow$ &  $10.3_{\uparrow}$ \\
                                         & \textit{Word}            & 6.3±0.80                                & 51.8±0.91                              & 12.9±0.26                               & 5.3±0.45                               & 18.7±0.28 \\
                                         & \textit{Word(\textbf{our})}   & 4.8±0.26  & \textbf{50.4±0.60}&  11.7±0.24 &   \textbf{5.1±0.58} &  17.4±0.66 \\                                        
                                         & \textit{RP}(\%)         & $23.8_{\uparrow} $   & $2.7_{\uparrow}$ & $9.3_{\uparrow}$ &  $3.8_{\uparrow}$ & $7.0_{\uparrow}$  \\                                        
 \midrule
\multirow{7}{*}{\small{$BERT_{base}$}}               & \textit{w/o}        & 2.6±0.18                                & 47.3±0.47                              &    6.9±0.21                            & 2.4±0.47                               & 11.5±1.19       \\
                                         & \textit{Sent}         & 2.2±0.24                                & 44.5±0.37                              &    6.3±0.29                            & 2.4±0.56                               & 11.3±1.44       \\
                                         & \textit{Sent(\textbf{our})}&  2.1±0.20              &  \textbf{44.3±0.54}             &\textbf{5.9±0.30}  &  2.3±0.49  &  11.2±1.31   \\
                                         & \textit{RP}(\%)           &  $4.5_{\uparrow}$             &  $0.4_{\uparrow}$              & $9.5_{\uparrow}$   & $4.2_{\uparrow}$  & $0.9_{\uparrow}$   \\
                                         & \textit{Word}            & 2.1±0.20                                & 45.6±0.37                              & 6.5±0.25                               & 2.3±0.54                               & 11.1±1.44       \\
                                         &  \textit{Word(\textbf{our})}  & \textbf{1.9±0.13}   &  45.5±0.37             &  6.4±0.23  &  \textbf{2.2±0.56}  & \textbf{10.8±1.29}   \\
                                         & \textit{RP}(\%)            & $9.5_{\uparrow}$    &  $0.2_{\uparrow}$            & $1.5_{\uparrow}$   & $4.3_{\uparrow}$   &$2.7_{\uparrow}$   \\
\bottomrule
\end{tabular}
\end{table*}
\subsection{Baselines and Settings}
Our \textit{AMP} is evaluated by integrating to two recent proposed \textit{Mixup} variants. We choose five popular sentence classification models as the backbone to test the performance of all \textit{Mixups} on the five benchmark datasets.

\textbf{Classification backbone.} We test \textit{Mixups} on five classification backbones. ${LSTM_{rand}}$ and $LSTM_{glove}$~\cite{wang2016attention} are two versions of bi-directional Long Short Term Memory(LSTM) with attention, where the former uses randomly initiated word embeddings and the latter uses GloVe~\cite{pennington2014glove} initiated word embeddings. $CNN_{rand}$ and $CNN_{glove}$~\cite{kim2014convolutional} are two versions of convolutional neural networks. They are fed with randomly and GloVe initiated word embeddings, repectively. The above four methods are popular sentence classification models without pre-training techniques. We employ $BERT_{base}$~\cite{devlin2018bert} as the pre-training classification backbone.

\textbf{Mixup}. We choose three popular \textit{Mixup} variants for sentence classification as baselines. \textit{\textbf{WordMixup}}~\cite{guo2019augmenting} is the straightforward application of \textit{Mixup} on NLP tasks where linear interpolation applying on the word embedding level (first layer). \textit{\textbf{SentMixup}}~\cite{verma2019manifold,sun2020mixup} is the \textit{Mixup} applying to NLP tasks where linear interpolation is conducted in the last layer of hidden states. \textit{\textbf{Non-linear Mixup}} is the non-linear version of \textit{SentMixup}. 

\textbf{AMP}. \textit{\textbf{WordAMP}} is applied on the word embedding level, the same as \textit{WordMixup}. \textit{\textbf{SentAMP}} is applied on the last layer of hidden states, the same as \textit{SentMixup}. 


We obtained the source codes of backbone models from the public available implementations\footnote{
LSTM: https://github.com/songyouwei/ABSA-PyTorch\\
CNN: https://github.com/harvardnlp/sent-conv-torch\\
BERT: https://github.com/huggingface/transformers\\
GloVe: https://nlp.stanford.edu/projects/glove/}. In our experiments, we follow the exact implementation and settings in ~\cite{kim2014convolutional,wang2016attention,devlin2018bert,guo2019augmenting,verma2019manifold}. Specifically, we use filter sizes of 3, 4, and 5, each with 100 feature maps; dropout rate of $0.5$ and L2 regularization of 1e-8 for the \textit{CNN} baselines. We use hidden size of $1024$ of single-layer; dropout rate of $0.5$ and L2 regularization of 1e-8 for the \textit{LSTM} baselines. For datasets without a standard development set, we randomly select 10\% of training data as a development set. Training is done through Adam~\cite{kingma2014adam} over mini-batches of size 50 (\textit{CNN}, \textit{LSTM}) and 24 ($BERT_{base}$) respectively. The learning rate is 2e-4 for \textit{CNN} and \textit{LSTM}, and 1e-5 for $BERT_{base}$. The word embeddings are 300 dimensions for \textit{CNN} and \textit{LSTM}. The step size $\varepsilon=0.002$ for all experiments. The $\alpha$ for all \textit{Mixup} is set to one. For each dataset, we train each model 10 times with different random seeds each with 8k steps and compute their mean error rates and standard deviations.

\subsection{Main results}
To evaluate the predictive performance of \textit{AMP}, we conduct five sets of experiments. For each setting, we compare the performance of without \textit{Mixup} (\textit{w/o}), \textit{WordMixup} (\textit{Word}), \textit{SentMixup} (\textit{Sent}) and \textit{non-linear Mixup}(\textit{non-linear}\footnote{Only on $CNN_{glove}$ our baseline results close to the baseline results reported in~\cite{guo2020nonlinear}). For a fair comparison, we only cite the results of $CNN_{glove}$ of non-linear \textit{Mixup}.}. As presented in Table~\ref{tab:2}, \textit{AMP} outperform \textit{Mixup} comparison baselines. For example, compared with the \textit{Sent} baseline over $\tiny CNN_{glove}$, \textit{Sent(our)} achieves a significant improvement on all five datasets. For instance,  \textit{Sent(our)} outperform \textit{Sent} on the TREC, SST2 and MR datasets over $\tiny CNN_{glove}$, the relative improvements are $31.3\%$, $8.6\%$ and $10.3\%$, respectively\footnote{Our methods are tuned on $CNN_{glove}$ may cause the significant higher level of improvements.}. Compared with \textit{Word} over $\tiny RNN_{glove}$, \textit{Word(our)} reduces the error rate over $1.2\%$ (up to $5.9\%$) on all five testing datasets. 
Interestingly, one can see that the \textit{Word(our)} outperform \textit{Non-linear Mixup} on three out of five datasets. That shows the slightly relaxing of LLC achieves similar sometimes even better results than changing the LLC into a non-linear version.

We use different initial embeddings to evaluate the effectiveness of augmentation as~\cite{guo2019augmenting}. From the embedding perspective, we have three kinds of embeddings: the randomly initiated embeddings ($\tiny{RNN_{rand}}$ and $\tiny{CNN_{rand}}$), the pre-trained fixed embeddings ($\tiny{RNN_{glove}}$ and $\tiny{CNN_{glove}}$) and the pre-trained context-aware embeddings ($\tiny BERT_{base}$).  For each kind of embeddings, \textit{AMP} outperforms the \textit{Mixup} baselines. For instance, when compared with \textit{Sent} under randomly initiated embeddings, the proposed method \textit{Sent(our)} obtains lower predictive error rate on eight out of ten experiments. While \textit{Word(our)} outperforms \textit{Word} on nine out of ten experiments. Similar results can be observed on the pre-trained embeddings settings. Even under the context-aware embeddings setting ($BERT_{base}$), our \textit{AMP} can further improve the performance against the \textit{Mixup} with advanced backbone models. Significantly, on SST1, our method help $BERT_{base}$ outperforms the SOTA model ($BERT_{large}$, $44.5$) ~\cite{munikar2019fine}, which is as two times large as $BERT_{base}$. The results show the effectiveness of our method.
\begin{table}[h]
\caption{The results of $BERT_{base}$ with \textit{SentAMP} on low-resource settings. The experiments are run ten times on each scaled TREC datasets. The average error rate and standard deviation are reported.\label{tab:6}}
\centering
\setlength{\tabcolsep}{2.0mm}
\begin{tabular}{cccccccc}
\hline
 \% & labels      & \textit{Sent}  & \textit{Sent(\textbf{our})} &\textit{RP(\%)} \\ \hline
3  & 160 & 51.0±7.34 & \underline{42.1±7.34} &\textbf{+17.5}  \\
4  & 215 & 29.8±4.05 & \underline{25.6±4.01} &+14.1    \\
5  & 270 & 10.2±1.00 & \underline{9.2±0.80} &+9.8   \\
10 & 543 & 5.1±0.64 & \underline{4.6±0.37} &+9.8   \\
15 & 815 & 4.1±0.64 & \underline{4.0±0.67} &+2.4   \\
20 & 1089 & 3.6±0.62 & \underline{3.5±0.48} &+2.8   \\
40 & 2179 & 2.9±0.35 & \underline{2.7±0.38} &+6.7   \\
80 & 4359 & 2.2±0.17 & \underline{2.1±0.10} &+4.5   \\
100 & 5452 & 2.2±0.24 & \underline{2.1±0.20} &+4.5  \\
 \hline
\end{tabular}
\end{table}
\subsection{Low-resource conditions}
With low resources, the under-fitting caused by the strict LLC has a serious impact on the model generalization. To evaluate our \textit{AMP} performance with different amounts of data, particularly in the case of low-resource settings. We scale the size of the dataset by a certain ratio of data for each category. If the scaled category is less than $0$, we retain at least one sample. We randomly generate ten different datasets for each scale ratio and then run the experiment on each dataset. The mean error rate and standard deviation are reported. As shown in Table~\ref{tab:6}, we can see that our method reduces the mean error rate against \textit{Mixup} with a significant margin. For instance, \textit{Sent(our)} reduces the error rate over \textit{Sent} with $17.5\%$ and $14.1\%$ on $3\%$ and $4\% $ training data, separately. AMP works well as we expected in low resource conditions for its effectiveness in relaxing LLC in \textit{Mixup}. 
\begin{table}[t]
\caption{Ablation study.\label{tab:3}}
\centering
\setlength{\tabcolsep}{2.6mm}
\begin{tabular}{cccc}
\hline
Method &Model                 & Operation       & TREC      \\ \hline
\multirow{8}{*}{\textit{Word}}&\multirow{4}{*}{\small{$CNN_{glove}$}}  & \textit{Baseline}     & 7.9±0.12 \\ 
                      && \textit{+RandOp}     & 6.3±0.80 \\ 
                      && \textit{+MaxOp}    & 4.7±0.35 \\
                      && \textit{AMP}    & 4.8±0.26 \\\cline{2-4}
&\multirow{4}{*}{\small{$BERT_{base}$}} & \textit{Baseline}     & 2.6±0.18 \\ 
                      && \textit{+RandOp}     & 2.1±0.24 \\ 
                      && \textit{+MaxOp}    & 2.0±0.23 \\
                      && \textit{AMP}    & 1.9±0.13 \\\hline\hline
\multirow{8}{*}{\textit{Sent}}&\multirow{4}{*}{\small{$CNN_{glove}$}}   & \textit{Baseline}     & 7.9±0.12 \\ 
                      && \textit{+RandOp}     & 6.7±0.23 \\ 
                      && \textit{+MaxOp}    & 4.8±0.22 \\
                      && \textit{AMP} & 4.6±0.33 \\\cline{2-4}
&\multirow{4}{*}{\small{$BERT_{base}$}} & \textit{Baseline}     & 2.6±0.18 \\ 
                      && \textit{+RandOp}     & 2.2±0.24 \\ 
                      && \textit{+MaxOp}    & 2.1±0.13 \\
                      && \textit{AMP} & 2.1±0.15 \\\hline
\end{tabular}
\end{table}
\begin{table}[t]
\caption{The results under different setting of $\alpha$.\label{tab:5}}
\centering
\setlength{\tabcolsep}{1.4mm}
\begin{tabular}{ccccc}
\hline
$\alpha$                  & Methods & TREC      & SST2      & MR        \\ \hline
\multirow{3}{*}{0.2} & \textit{Word}      & 1.9±0.13 & 6.3±0.23 & 11.0±1.25 \\
                       & \textit{Word(our)}         & 1.8±0.13 & 6.0±0.20 & 10.9±1.22 \\ 
                       & \textit{RP}(\%)         & +5.3 &      +4.8 & +0.9 \\ \hline
\multirow{3}{*}{0.5} & \textit{Word}     & 1.9±0.13 & 6.7±0.24 & 11.1±1.25 \\
                       & \textit{Word(our)}         & 1.9±0.16 & 6.1±0.18 & 10.8±1.25 \\ 
                       & \textit{RP}(\%)        & +0.0 & \textbf{+8.9} &+2.7 \\ \hline
\multirow{3}{*}{1.0} & \textit{Word}     & 2.1±0.20 & 6.5±0.25 & 11.1±1.44 \\
                       & \textit{Word(our)}         &2.0±0.12 & 6.4±0.23 & 10.8±1.29 \\ 
                       & \textit{RP}(\%)         & +4.8 & +1.5 & +2.7 \\ \hline
\multirow{3}{*}{1.5} & \textit{Word}     & 2.1±0.18 & 6.8±0.13 & 11.2±1.44 \\
                       & \textit{Word(our)}         & 2.0±0.12 & 6.5±0.28 & 11.0±1.34 \\
                       & \textit{RP}(\%)         & +4.8 & +4.4 & +1.8 \\ \hline
\end{tabular}
\end{table}
\subsection{Ablation study}
To further understand the Max Operation (MaxOp) and Min Operation (MinOp) effects in \textit{AMP}, we make several variations of our model. The variations are tested under $CNN_{glove}$ and $BERT_{base}$ on TREC. As presented in Table~\ref{tab:3}, the model trained without augmentation is denoted as $Baseline$. $+RandOp$ is identical to the model trained with \textit{Mixup}, $+MaxOp$ indicates \textit{Mixup} with MaxOp is used for model training, \textit{AMP} is the fully functional method of our proposed method. As the results presented in Table~\ref{tab:3}, MaxOp contributes the majority cut down of error rate. For instance, the $CNN_{glove}$ under \textit{Sent Mixup} settings, MaxOp reduces the error rate from $6.7$ to $4.8$. That suggests the effectiveness of adversarial perturbation in relaxing the LLC in \textit{Mixup}. The comparison in MinOp can mostly (three out of four times) further reduce the error rate. Specifically, it brings down the mean error rate from $4.8$ to $4.6$ on $CNN_{glove}$. That indicates the effectiveness of MinOp in eliminating the influence of step size.


\subsection{Mix ratio distribution}
To analyze the effects of different shapes of mixing coefficient distributions, we compare \textit{Word(out)} with \textit{Word} on $BERT_{base}$ on four $\alpha$ settings (from $0.2$ to $1.5$) and three datasets: TREC, SST2, and MR. The $\alpha$ is the parameter of the $Beta$ distribution. It controls the shape of how the mixing coefficient $\lambda$ is distributed.  As presented in Table~\ref{tab:5}, our method can achieve lower mean error rates than \textit{Word} on all $\alpha$ settings. For instance, \textit{Word(our)} achieve $8.9\%$ lower mean error rate than \textit{Word} on SST2 with $\alpha=0.5$. The improvements come mainly from training the models with the slightly non-linear data generated by AMP. 

\begin{figure}[t]
\centering
\subfigure[Random pair1]{
\includegraphics[width=0.43\columnwidth]{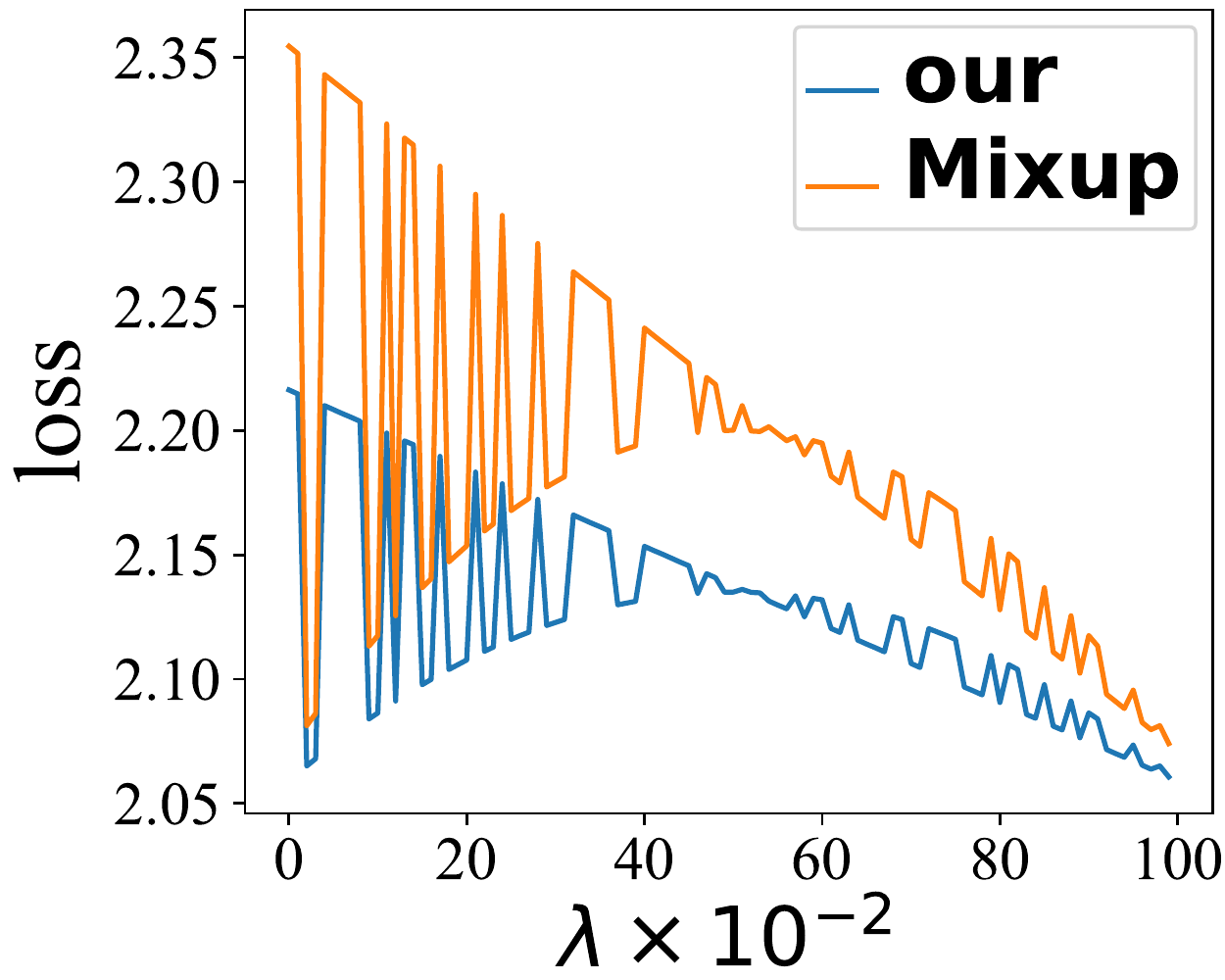}
\label{fig:fig3_1}
}
\subfigure[Random pair2]{
\includegraphics[width=0.43\columnwidth]{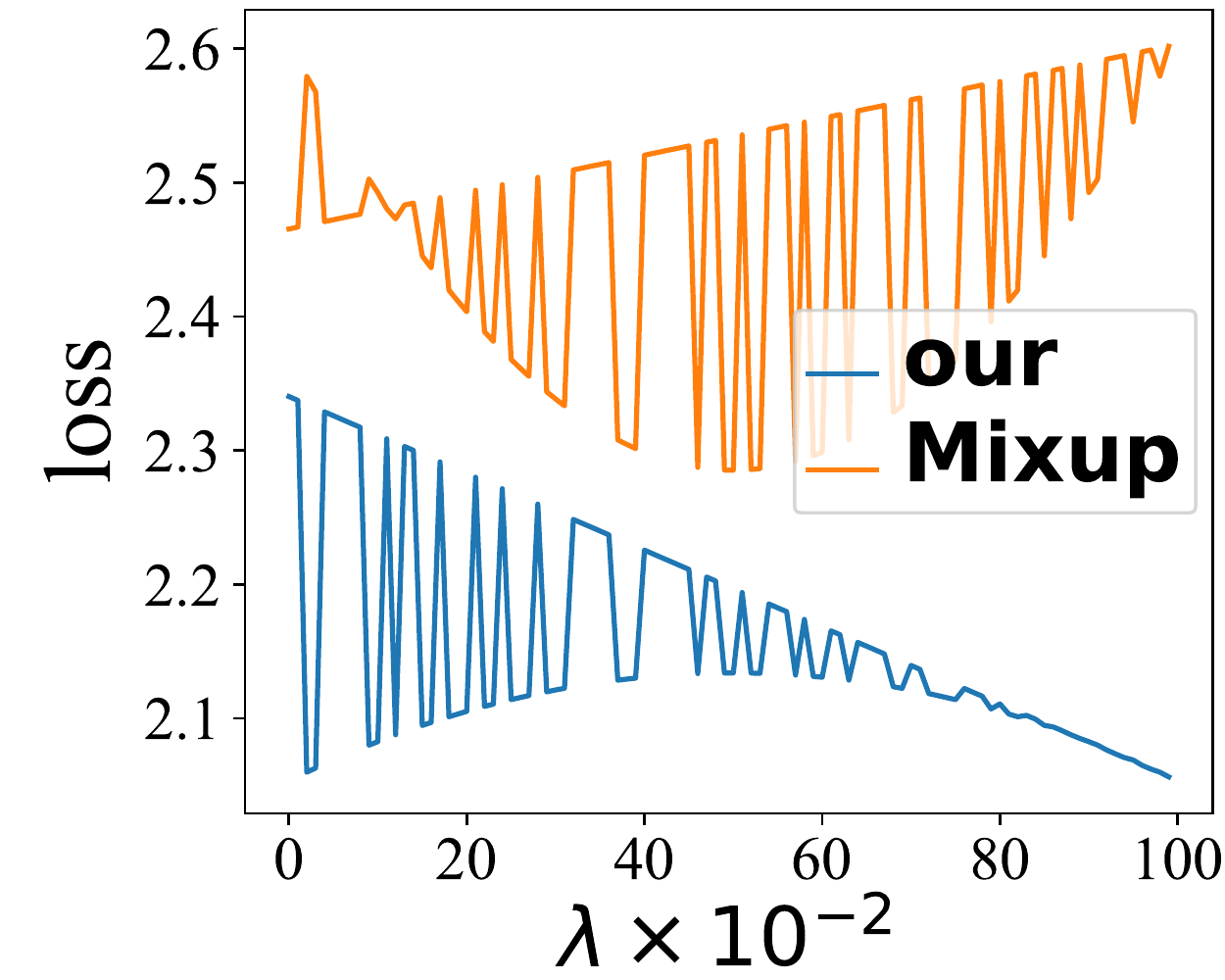}
\label{fig:fig3_2}
}
\subfigure[Full-size testing set]{
\includegraphics[width=0.9\columnwidth]{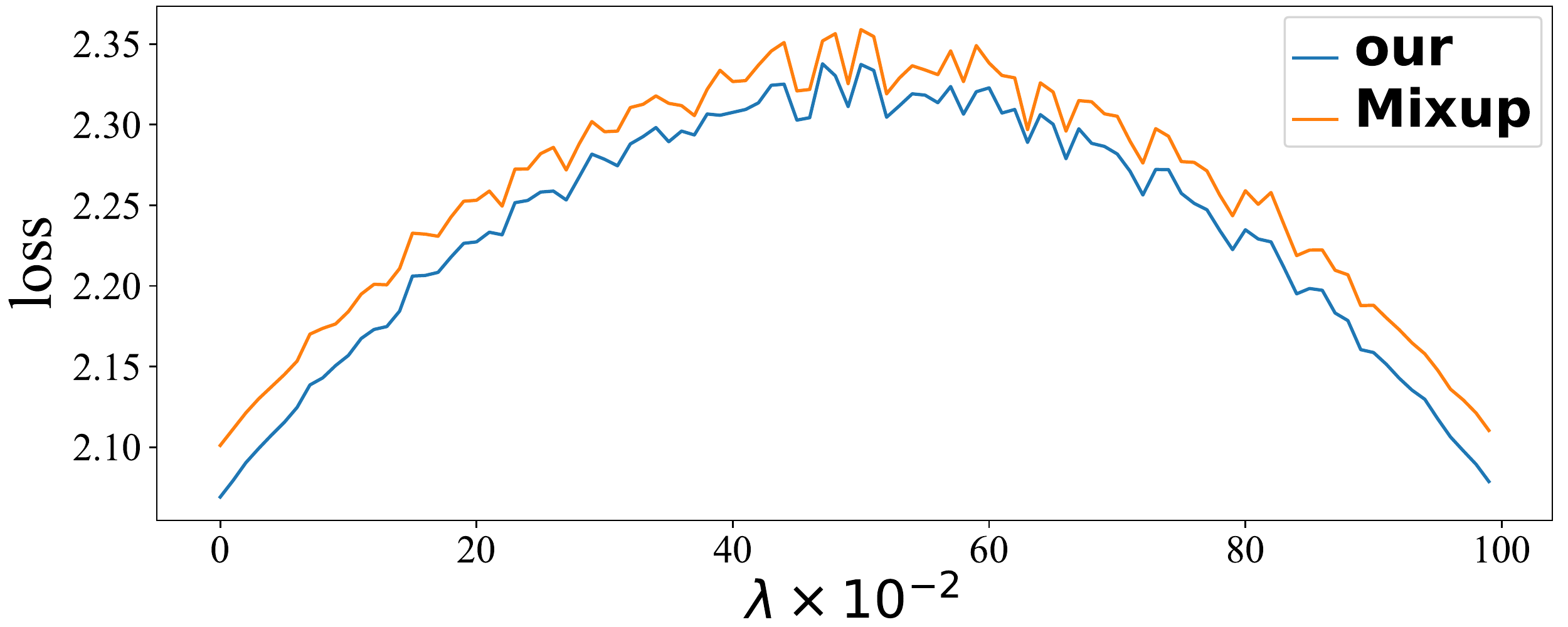}
\label{fig:fig3_3}
}
\caption{The visualization of loss on unseen synthetic data. The results conduct by $BERT_{base}$ on $3$\% TREC dataset, as listed in Table~\ref{tab:6}.\label{fig3}}
\end{figure}
\subsection{Visualization}
To intuitively demonstrate the effects of relaxing LLC,  we visualize the loss of networks trained by our \textit{AMP} and \textit{Mixup}. The synthetic data is generated strictly follow the LLC based on the testing data. The network trained with relaxed LLC has a smaller loss value shows the effectiveness of our method in alleviate under-fitting. As shown in Figure~\ref{fig:fig3_1},~\ref{fig:fig3_2} and~\ref{fig:fig3_3}, we draw the losses on synthetic data generated with mixing coefficient $\in[0,1]$. Figure~\ref{fig:fig3_1} and~\ref{fig:fig3_2} each uses one random pair of data in the testing set for generating. For two random pair $(x_1,y_1)(x_4,y_4)$ and $(x_2,y_2)(x_3,y_3)$, we calculate the Mixup loss of each pair on different $\lambda$ to get Figure~\ref{fig:fig3_1} and~\ref{fig:fig3_2}. The loss curves on random pairs are not symmetric for the loss of each example of the pairs are different. The loss curves are encouraged (by LLC) to be a line in-between two examples. The line should start with the loss of one example and end with the loss of another example. The Mixup loss (interpolation on cross-entropy loss) and the different examples result in different shapes of the loss curves in Figure~\ref{fig:fig3_1} and~\ref{fig:fig3_2}.As illustrated in Figure~\ref{fig:fig3_1} and~\ref{fig:fig3_2}, one can observe that \textit{AMP} have a smaller loss than \textit{Mixup}. That indicates the effectiveness of training on the slightly non-linear synthetic data in the micro view. 

Figure~\ref{fig:fig3_3} uses the full-size testing set for generating. Figure~\ref{fig:fig3_3} shows the average loss over all synthetic data generated with the full-size testing set. We freeze the random seeds; thus, we can freeze the data pairs. Let the testing dataset be $X=[(x_1,y_1),(x_2,y_2),(x_3,y_3),(x_4,y_4)]$. The synthetic data is generated by $\lambda X+(1-\lambda)X$, where $X'=[(x_4,y_4),(x_3,y_3),(x_2,y_2),(x_1,y_1)]$ is shuffled $X$. So, the loss when $\lambda=0$ and $\lambda=1$ are identical. Similarly, we can get a symmetric picture as Figure~\ref{fig:fig3_3}.One can observe that our method can achieve a significantly smaller average loss than \textit{Mixup} in the macro view. The visualizations verified our assumption that relaxing LLC can further regularize models.

\section{Related work}
\textbf{\textit{Mixup} on text classification}. Text classification has achieved remarkable improvements underlying some effective paradigms, e.g., \textit{CNN}~\cite{kim2014convolutional}, attention-based \textit{LSTM}s~\cite{wang2016attention}, GloVe~\cite{pennington2014glove} and  \textit{BERT}~\cite{devlin2018bert}, etc. The large scale parameter of the model tends to generalize poorly in low-resource conditions. To overcome the limitation, \textit{Mixup}~\cite{zhang2017mixup} is proposed as a data augmentation based regularizer. Few researches explore the  \textit{Mixup}~\cite{guo2019mixup,zhang2020seqmix,guo2020nonlinear} on NLP tasks. For classification, ~\cite{guo2019augmenting} suggest applying \textit{Mixup} on particular level of networks, i.e., word or sentence level. Although these work make promising progress, the mechanism of Mixup is still need to be explored. 

\noindent\textbf{Adversarial Training.} The min-max formulation of adversarial training has been theoretically and empirically verified~\cite{beckham2019adversarial,xu2020adversarial,pang2019mixup,archambault2019mixup,lee2020adversarial,miyato2015distributional,miyato2018virtual,miyato2016adversarial}. Such training procedure first generates adversarial examples that might maximize the training loss and then minimizes the training loss after adding the adversarial examples into the training set~\cite{madry2017towards}. The Fast Gradient Sign Method (FGSM)~\cite{goodfellow2014explaining} is an efficient one-step method.  
Inspired by the min-max formulation of adversarial learning, we organize our method into a min-max-rand formulation.

\section{Conclusion}
For relaxing Locally Linear Constraints (LLC) in \textit{Mixup} to alleviate the under-fitting, this paper proposes an Adversarial Mixing Policy (\textit{\textbf{AMP}}). Inspired by the adversarial training, we organize our method into a \textit{\textbf{min-max-rand}} formulation. The proposed method injects slightly non-linearity in-between synthetic examples and synthetic labels without extra parameters. By training on these data, the networks can compatible with some ambiguous data and thus reduce under-fitting. Thus, the network will be further regularized to reach better performance. We evaluate our method on five popular classification models on five publicly available text datasets. Extensive experimental results show that our \textit{AMP} can achieve a significantly lower error rate than vanilla \textit{Mixup} (up to 31.3\%), especially in low-resource conditions(up to 17.5\%). 
\section{Acknowledgments}
We thank Prof.Xiaojie Wang and Prof.Fangxiang Feng from BUPT for their valuable feedback on an earlier draft of this paper, and Yang Du from XDF for her suggestions of English writing for the final revision. We also thank anonymous reviewers for their helpful comments. 
\bibliographystyle{acl_natbib}
\bibliography{emnlp2021}


\end{document}